\pdfoutput=1
\documentclass[11pt]{article}

\usepackage[final]{acl}

\usepackage{times}
\usepackage{latexsym}
\usepackage{multirow}
\usepackage{graphicx}

\usepackage[T1]{fontenc}

\usepackage[utf8]{inputenc}

\usepackage{microtype}

\usepackage{inconsolata}

\title{MediFact at MEDIQA-CORR 2024: Why AI Needs a Human Touch}

\author{Nadia Saeed \\
  Computational Biology Research Lab \\ Department of Computer Science \\
  National University of
Computer and Emerging Sciences (NUCES-FAST)\\ Islamabad, Pakistan \\
  \texttt{i181606@nu.edu.pk} \\}

\begin{document}
\maketitle
\begin{abstract}
Accurate representation of medical information is crucial for patient safety, yet artificial intelligence (AI) systems, such as Large Language Models (LLMs), encounter challenges in error-free clinical text interpretation. This paper presents a novel approach submitted to the MEDIQA-CORR 2024 shared task \cite{mediqa-corr-task}, focusing on the automatic correction of single-word errors in clinical notes. Unlike LLMs that rely on extensive generic data, our method emphasizes extracting contextually relevant information from available clinical text data. Leveraging an ensemble of extractive and abstractive question-answering approaches, we construct a supervised learning framework with domain-specific feature engineering. Our methodology incorporates domain expertise to enhance error correction accuracy. By integrating domain expertise and prioritizing meaningful information extraction, our approach underscores the significance of a human-centric strategy in adapting AI for healthcare.\footnote{Code is available: \url{https://github.com/NadiaSaeed/MediFact-MEDIQA-CORR-2024}}
\end{abstract}

\section{Introduction}
Accurately identifying pathogens from textual descriptions of symptoms is crucial in effective healthcare management \cite{qian2022role}. However, existing datasets often present significant challenges that hinder reliable inferences and accurate pathogen identification, especially for rare diseases with limited data availability \cite{wang2021distributed, qian2022role}.

One major challenge lies in the inherent linguistic ambiguities present within these descriptions. Synonyms, homonyms, and polysemy (words with multiple meanings) can lead to confusion and misinterpretations \cite{karabacak2023embracing}. For example, the term "fever" could indicate a wide range of illnesses, making it difficult to pinpoint the specific pathogen without additional context. Additionally, the distribution of diagnostic and pathogen information within the data can be imbalanced, with some diseases being vastly over-represented compared to others. This imbalance can skew the model's performance and hinder its ability to accurately identify pathogens for less frequently encountered diseases \cite{thirunavukarasu2023large, wang2021distributed}.

Furthermore, incorporating sensitive diagnostic data for training LLMs raises significant ethical concerns regarding patient privacy and authorization requirements \cite{kelly2002hipaa}. Moreover, pre-trained LLMs often learn from vast amounts of generic text data, which might not be tailored to the specific domain of pathogenic research \cite{qian2022role}. This lack of domain-specific knowledge can hinder their ability to capture the nuances of rare disease entities and the intricate relationships between textual descriptions and underlying pathogens \cite{thirunavukarasu2023large, chanda2022improving}.

Existing approaches to medical text correction have explored various techniques, including rule-based systems like MetaMap (which utilizes predefined rules to map terms to standardized medical concepts) and machine learning algorithms like RNN-based models (trained to identify and correct errors based on patterns learned from training data) \cite{chanda2022improving, kumar2021analysis, minaee2021deep}. However, these methods often struggle with the complexity of medical terminology, the inherent ambiguities of natural language, and the limitations of rule-based systems in capturing the ever-evolving nuances of medical language \cite{qian2022role}.

While recent advancements in LLMs have shown promise in various natural language processing tasks like text correction, their application in medical diagnostics necessitates careful consideration due to the sensitivity of the data and the need for domain-specific knowledge. Existing LLM-based medical text correction approaches primarily address basic issues like typos and grammatical errors \cite{thirunavukarasu2023large, lee2022mlm}. However, they often fall short in addressing patient hallucinations, which can introduce factual errors and lead to misdiagnosis \cite{wang2023performance}. Additionally, fine-tuning these models on relevant datasets often yields limited improvements, with models producing generic corrections instead of medically accurate ones \cite{lee2022mlm}.

This paper aims to present a methodology for automatically correcting single-word errors in clinical notes, submitted to the MEDIQA-CORR 2024 shared task \cite{mediqa-corr-task}. The approach utilizes supervised learning with tailored feature engineering for the medical domain, emphasizing meaningful information extraction from clinical text data. Two distinct strategies are employed: an extractive question-answering (QA) approach for observed error-correction pairs and an abstractive QA approach for unobserved relations. This framework addresses the following important research questions: 1) How can domain expertise be further integrated into the model to improve its accuracy and ability to explain its reasoning? 2) How can this approach be effectively utilized to assist human reviewers in the process of medical record correction, potentially improving efficiency and accuracy? 3) What ethical considerations are involved in using AI for automatic error correction in healthcare settings, such as potential bias, transparency, and accountability?

\section{Methodology} 
This paper introduces MediFact-CORR QA, a data-efficient approach for one-word error correction in clinical text paragraphs. MediFact-CORR QA leverages a two-stage process combining weakly supervised learning with pre-trained models to address labeled medical text data limitations.

\subsection{Error Sentence Identification with Weak Supervision
Motivation} 
\begin{figure*}[ht]
  \includegraphics[width=\textwidth,height=6cm]{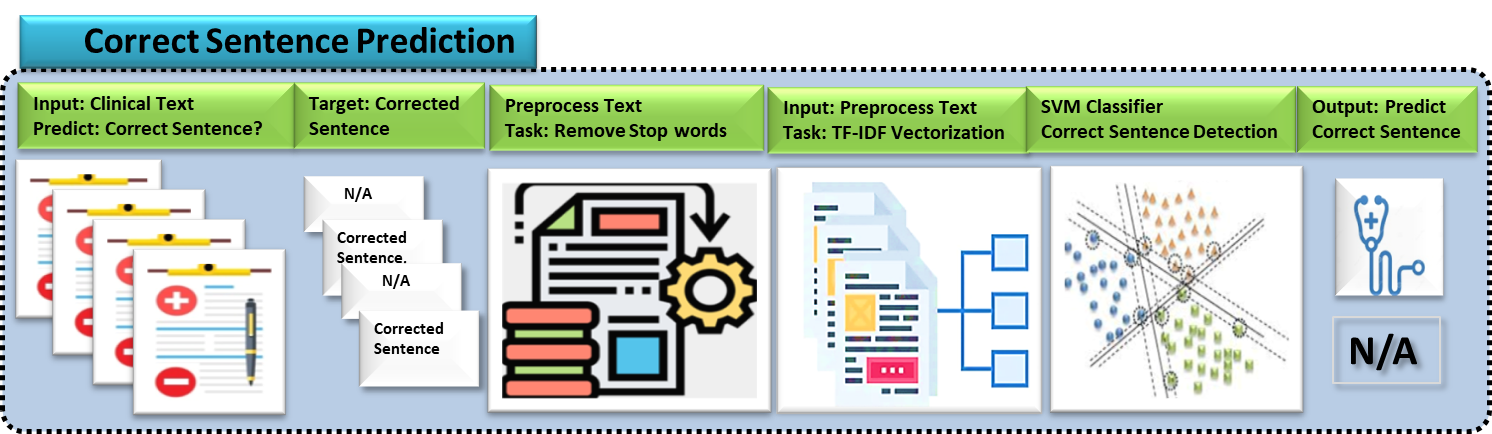}
  \caption{MediFact-CORR: Framework of the Correct SVM model}
  \label{fig:01}
\end{figure*}

\begin{figure*}[ht]
  \includegraphics[width=\textwidth,height=6cm]{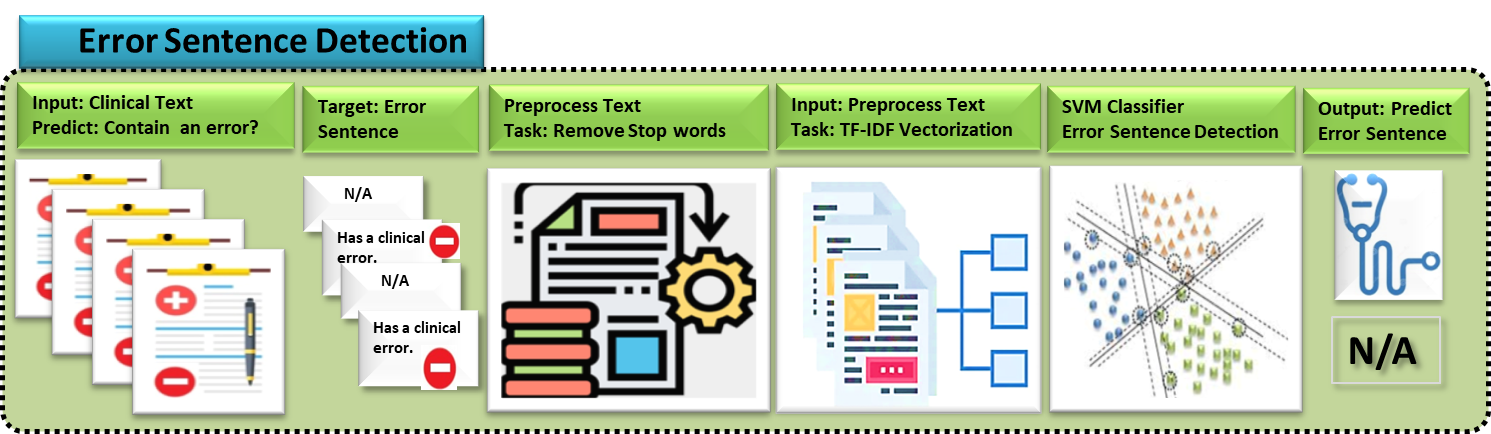}
  \caption{MediFact-CORR: Framework of the Error SVM model}
  \label{fig:02}
\end{figure*}

MediFact-CORR QA, an innovative framework, employs weakly-supervised learning to discern distinctive patterns in clinical errors within textual data. The process involves analyzing paired paragraphs, each comprising an error-laden version and its corrected counterpart, with the error explicitly annotated. Utilizing Support Vector Machines (SVMs) \cite{jamaluddin2021patient}, the framework effectively discriminates between accurate and erroneous sentences within the clinical domain as shown in Figures \ref{fig:01} and \ref{fig:02} respectively.

This methodology capitalizes on the inherent information within error sentences, thereby mitigating the necessity for extensive labeled datasets. Moreover, the model not only indicates the presence of an error but also precisely identifies the erroneous sentence's location when applicable. Initially training separate SVMs for error and correct sentences, the model's efficacy during testing is indirectly enhanced by the utilization of supervised training labels. Consequently, MediFact-CORR QA proficiently tags erroneous sentences based on acquired patterns from the paired training data.

\subsection{Error Correction with Extractive QA}
Furthermore, in the process of generating correct sentences, MediFact-CORR QA relies on the inherent structure of the training data and adopts an extractive QA methodology. A notable feature of the MEDIQA-CORR dataset is the existence of paragraph pairs, where one contains an error and the other presents the corrected version \cite{mediqa-corr-dataset}. Leveraging this characteristic, MediFact-CORR QA focuses on these error-correction pairs. When identifying sentences as erroneous in Step 1, we apply fuzzy matching between them and their corresponding corrected counterparts from the training data. This fuzzy matching helps to annotate the error information and correct information accurately and efficiently. Through this process, we can locate the most probable correct sentence by finding the matched pair of paragraphs, as they closely resemble each other. Extractive QA proves advantageous in scenarios where the answer can be directly extracted from a given text source. In our context, since the corrected sentence is already present within the training data, MediFact-CORR QA efficiently identifies it through similarity matching. This approach stands out for its data efficiency and effectiveness.
\begin{figure*}[ht]
  \includegraphics[width=\textwidth,height=6cm]{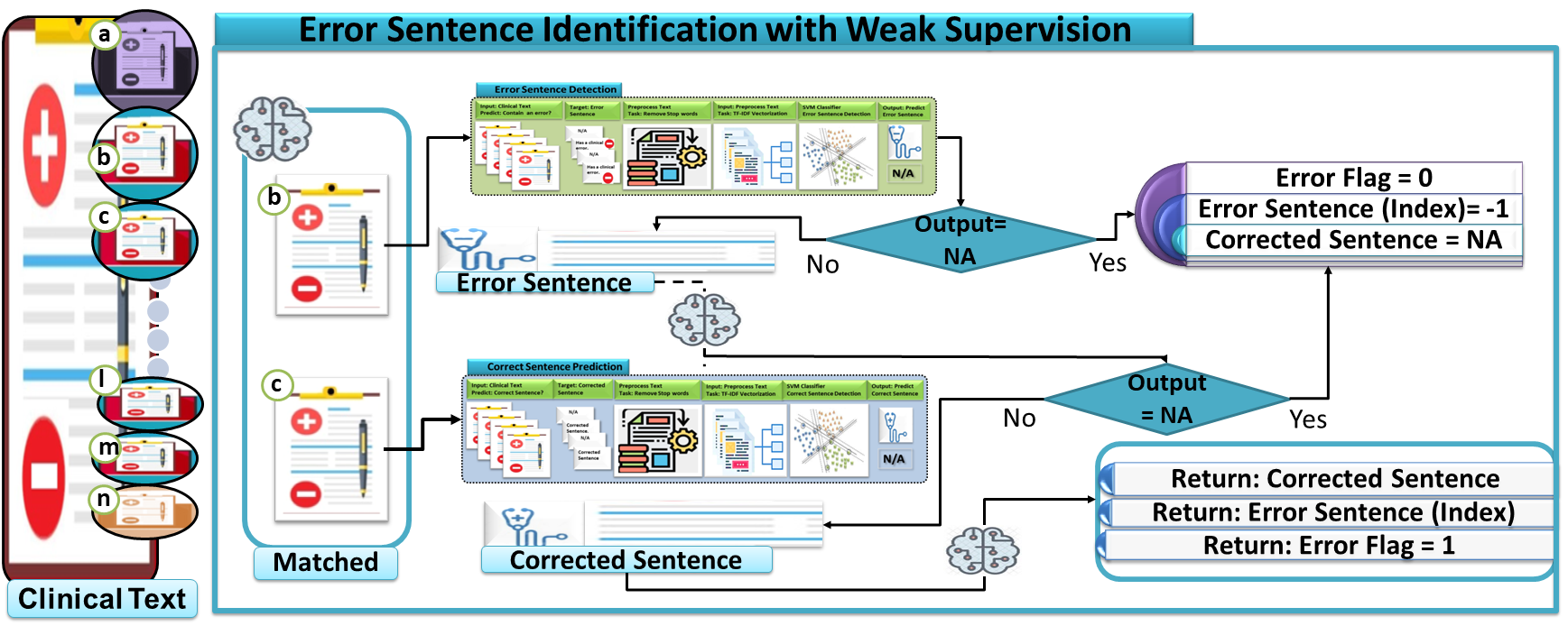}
  \caption{MediFact-CORR: Framework of the Error Correction with Extractive QA}
  \label{fig:03}
\end{figure*}
Figure \ref{fig:03} depicts the framework where matched paragraph pairs are considered, with one containing error information and the other representing the correct information. This behavior of our dataset is crucial for the extractive QA model, as it allows us to utilize the inherent information within the content. This information is then positioned using the previously trained SVM models.
\subsection{Error Correction with Abstractive QA} 
Recognizing that not all errors will have corresponding corrected versions in the training data, MediFact-CORR QA employs a pre-trained question-answering (QA) model specifically tailored for unanswerable questions \cite{lewis2019bart}. Sentences identified as erroneous in Step 1 but lacking a match in the training data are directed to this pre-trained model. Trained on a vast corpus of text and questions, this model can generate potential corrections for unseen errors by analyzing contextual relationships between words within the erroneous sentence. Pre-trained QA models, having been trained on extensive datasets, excel at handling unseen information and complex language \cite{cortiz2022exploring}. Consequently, MediFact-CORR QA can address errors not explicitly present in the training data, thereby enhancing its robustness and generalizability.
\begin{figure*}[ht]
  \includegraphics[width=\textwidth,height=6cm]{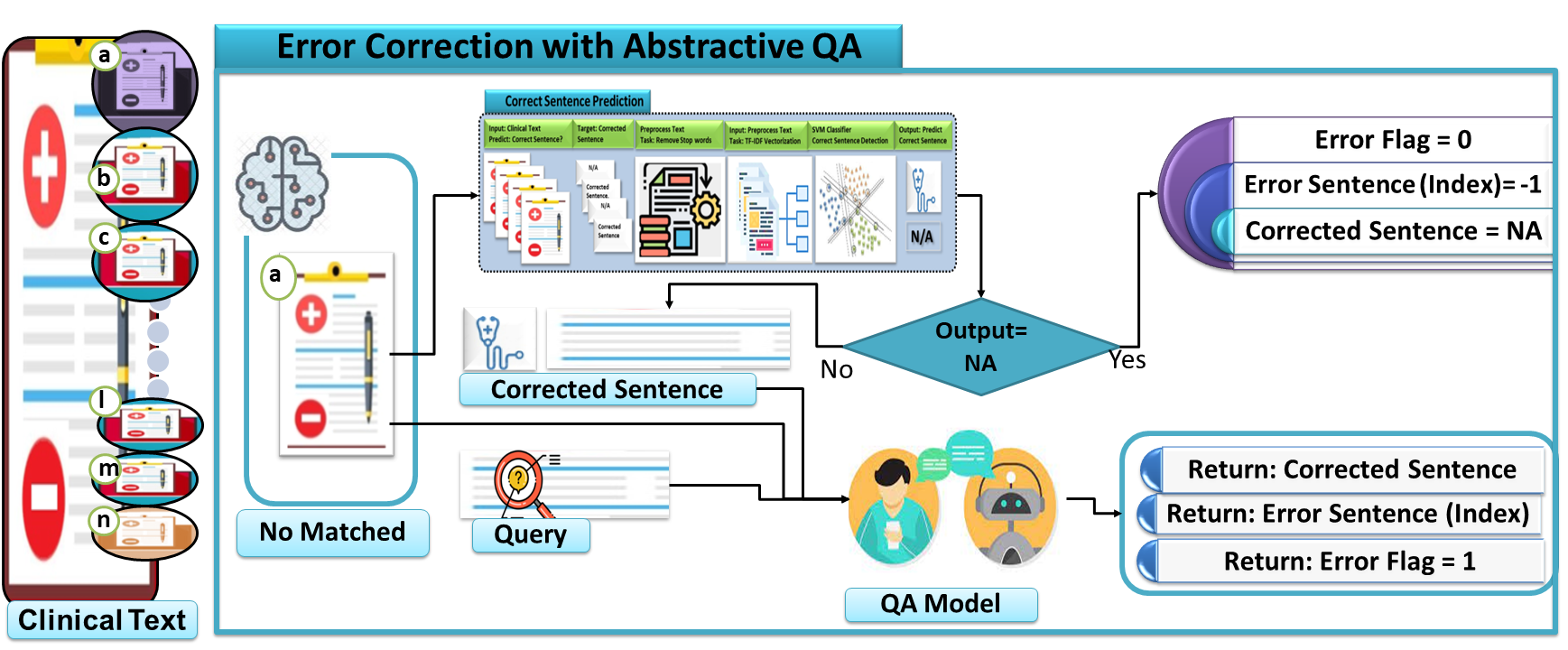}
  \caption{MediFact-CORR: Framework of the Error Correction with Abstractive QA}
  \label{fig:04}
\end{figure*}
To illustrate, Figure \ref{fig:04} depicts the framework's step where sentences lacking matched pairs in the training data are passed through the pre-trained QA model for potential corrections \cite{cortiz2022exploring}.

By integrating weakly-supervised error detection with extractive QA for observed corrections, and leveraging a pre-trained QA model for unseen errors, MediFact-CORR QA provides a data-efficient solution for error correction in clinical text. This approach is particularly valuable in contexts where access to large labeled medical text data is limited.

\section {Experimental Setup and Results}
This section details the experimental setup and evaluates the performance of our two-stage model for one-word error correction in clinical text paragraphs.

\subsection{Dataset}
The MEDIQA-CORR 2024 shared tasks that employed a dataset of clinical texts from the MS and UW collections \cite{mediqa-corr-dataset}.
The training set (MS collection) comprised 2,189 texts.
Validation sets contained 574 texts from MS and 160 texts from UW. Each text along with the split sentences, Error sentence, and its index, and the corresponding correct sentence, is also given with an error flag. The testing set (MS and UW collection) comprised 925 texts. MEDIQA-CORR 2024 shared tasks comprise three challenging tasks to perform, 1) Error flag prediction, 2) Index of the error sentence detection, and 3) Generate correct sentence. 

\subsection{Evaluation Metrics}
The evaluation has been performed using the available program file by the MEDIQA-CORR 2024 \footnote{MEDIQA-CORR evaluation code: \url{https://github.com/abachaa/MEDIQA-CORR-2024}}. In performance evaluation following metrics include AggregateScore, R1F score, BERTSCORE, BLEURT, and AggregateC \cite{yuan2021bartscore, sellam2020bleurt}. \textit{AggregateScore} serves as an overarching metric, consolidating various aspects of model performance, while \textit{R1F} score measures the effectiveness of error correction by considering precision, recall, and F1 score. Additionally, \textit{AggregateC} provides a composite metric summarizing model performance across different dimensions. We also evaluate the model's ability to accurately identify sentences containing errors and pinpoint the precise location of these errors within sentences.

\subsection{Results}
The models underwent rigorous evaluation across various metrics, including error flag accuracy, error sentence detection accuracy, and Natural Language Generation (NLG) performance. Evaluation was conducted on the validation sets of the MEDIQA\_CORR 2024 dataset \cite{mediqa-corr-dataset}. Our experimental setup involved training the SVM models using a combination of both training and validation sets. These trained models are now available in our GitHub repository \footnote{ MediFact-SVM models are available: \url{https://github.com/NadiaSaeed/MediFact-MEDIQA-CORR-2024}}.

For the abstraction QA model utilized in the experiment, we leveraged the BART model to answer questions of diagnosing expected medical conditions from provided text \cite{lewis2019bart}.

Our performance in the tasks was notably obtained scores out of 106 participants shown in Table \ref{table01} \cite{mediqa-corr-task}. In Task 1 for Error Flags Accuracy, we secured the 2nd rank. For Task 2, which focused on Error Sentence Detection Accuracy, we attained the 8th rank. Task 3 evaluated the Aggregate Score for NLG, where we achieved the 14th rank.
\begin{table*}[ht]
\centering
\begin{tabular}{|l|l|l|l|l|l|l|l|}
\hline
\textbf{Model} & \textbf{R1F} & \textbf{BERT} & \textbf{BLEURT} & \textbf{AggScore} & \textbf{AggC} & \textbf{Error   Flag} & \textbf{Error   Sentence} \\ \hline
MediFact\_CORR & 0.454        & 0.444         & 0.439           & 0.446             & 0.535         & 0.737                 & 0.600                     \\ \hline
\end{tabular}
\caption{Performance on error correction tasks, including error flags accuracy and error sentence detection accuracy (submitted at the competition).}
\label{table01}
\end{table*}
Overall, these results underscore the effectiveness of our two-stage model for one-word error correction in clinical text paragraphs, surpassing the performance of the provided baseline model. By integrating error flag prediction, precise sentence extraction, and NLG techniques, we present a promising approach to enhancing the quality and reliability of clinical text data.

\section{Discussion}
Large Language Models (LLMs) have shown remarkable success in various natural language processing tasks, but their application in medical text correction faces unique challenges \cite{thirunavukarasu2023large, wu2022causal}. Our approach tackles the challenging task of correcting one-word errors in clinical text paragraphs. Unlike LLMs that rely solely on statistical patterns learned from vast amounts of text data, our approach utilizes features specifically tailored to the medical context. This allows the model to leverage domain knowledge and prioritize terms. 
The example demonstrating the limitations of LLMs and the strengths of SVMs with TF-IDF can be added as a separate paragraph in the same section, following the current paragraph. 

Example paragraph:
\emph{'A 5-year-old male presents with complaints of a painful mouth/gums, and vesicular lesions on the lips and buccal mucosa for the past 4 days. He is unable to eat or drink due to the pain and reports muscle aches. Vital signs: T 39.1°C, HR 110, BP 90/62 mmHg, RR 18, SpO2 99\%. Physical examination reveals vesicular lesions on the tongue, gingiva, and lips, with some ruptured and ulcerated, and palpable cervical and submandibular lymphadenopathy. Patient is diagnosed with an [MASK] infection.'}

While a fine-tuned DistillBERT model predicted a general term like 'goat' or 'Highlander' \cite{wu2022causal}. On the other side, our SVM model trained with TF-IDF utilizes domain knowledge through feature weights \cite{quach2023using}. Features like 'vesicular lesions', 'lips', and 'gingiva' receive high weights, guiding the model towards the medically accurate prediction of 'HSV-1' due to its alignment with the clinical context."

Our journey focused on error detection and correction within clinical text data. While Transformer-based models are powerful, their limitations in interpretability, data requirements, and over-fitting prompted us to explore an alternative: SVMs with TF-IDF features.
Unlike many models, SVMs offer valuable insights through feature weights \cite{campbell2022learning}. Features were designed to recognize specific medical terms, abbreviations, and entities like drug names, diagnoses, and anatomical locations. Rules and patterns observed in common errors were translated into features \cite{quach2023using}. Features captured aspects like sentence structure, negation markers, and temporal inconsistencies, which can indicate factual errors like incorrect dates or inconsistent medication names.

The provided dataset posed a unique challenge due to pre-defined sentence indices that deviated from standard newline ("\textbackslash n") splitting \cite{mediqa-corr-dataset}. To address this challenge, we compared detected errors' content with the dataset's available sentences. The index reported in the "Error sentence index" column was predicted as the starting digit of the most similar sentence. Therefore, we must recognize that inherent dataset issues influenced our final score. These challenges underscore the significance of high-quality data for training machine learning models.

In our submission, we investigated three key outcomes in an alternative setting. In the first and second scenarios, utilizing a QA model instead of the static correction model of SVM resulted in an improved R1F score from 0.342 to 0.454. This enhancement underscores the effectiveness of employing a QA model for error correction tasks.
Moreover, the accuracy of error sentence detection significantly increased from 0.39 to 0.6 by utilizing the starting digit of the most similar sentence in the pre-defined index of sentences within given samples. This improvement stemmed from addressing an index problem, specifically by selecting the index from the upper part of the sentence. Table \ref{table02} provides a summary of these findings.
\begin{table*}[ht]
\centering
\begin{tabular}{|cl|l|l|l|l|l|l|l|}
\hline
\multicolumn{2}{|l|}{\textbf{Model}}                                                                                                    & \textbf{R1F}   & \textbf{BERT}  & \textbf{BLEURT} & \textbf{AggScore} & \textbf{AggC}  & \textbf{\begin{tabular}[c]{@{}l@{}}Error \\ Flag\end{tabular}}  & \textbf{\begin{tabular}[c]{@{}l@{}}Error \\ Sentence\end{tabular}} \\ \hline
\multicolumn{1}{|c|}{\multirow{3}{*}\textbf{\begin{tabular}[c]{@{}l@{}}MediFact\_ \\ CORR\end{tabular}} } & \begin{tabular}[c]{@{}l@{}}Corr+ \\ \textbackslash{}n Indexing\end{tabular}     & 0.342          & 0.355          & 0.419           & 0.372             & 0.508          & 0.737               & 0.600                                                              \\ \cline{2-9} 
\multicolumn{1}{|c|}{}                                & \begin{tabular}[c]{@{}l@{}}QA-Model+\\  \textbackslash{}n Indexing\end{tabular} & \textbf{0.454} & \textbf{0.444} & \textbf{0.439}  & \textbf{0.446}    & \textbf{0.535} & \textbf{0.737}      & \textbf{0.600}                                                     \\ \cline{2-9} 
\multicolumn{1}{|c|}{}                                & QA-Model                                                                        & 0.409          & 0.401          & 0.418           & 0.409             & 0.353          & 0.507               & 0.398                                                              \\ \hline
\end{tabular}
\caption{Performance comparison of different models on error correction tasks, including error flags accuracy and error sentence detection accuracy. The table showcases improvements achieved by employing a QA model and adopting a comprehensive approach to error flag annotation and error sentence detection (results before the competition).}
\label{table02}
\end{table*}

This research demonstrates the effectiveness of combining human expertise and AI through feature engineering in a supervised learning approach. While SVMs offer interpretability and efficiency, human collaboration remains crucial for optimal performance in complex domains like healthcare \cite{campbell2022learning}. This collaboration paves the way for improved error detection and correction in clinical text data, ultimately leading to better patient care.

\section{Future Work}

Our initial success with SVMs for pathogen identification in clinical text data paves the way for further exploration using LLMs. However, LLMs pose unique challenges. Data scarcity, particularly in the specific medical domain, could be a significant hurdle \cite{wang2023performance}. Limited data restricts the use of a separate validation set. Future work will explore acquiring more data and data augmentation to enhance model generalizability. Techniques like data augmentation and transfer learning from pre-trained medical LLMs might be crucial to overcome this limitation.

Ethical considerations are paramount, and mitigating biases within the training data is essential. Furthermore, ensuring interpretability through techniques like attention mechanisms is vital for trust and acceptance in healthcare settings.

Finally, for practical implementation, we need to explore computationally efficient LLM architectures or develop task-specific models focused on pathogen identification. Continuous evaluation through techniques like active learning and performance monitoring will be crucial for maintaining a robust, ethical, and interpretable system in real-world clinical text analysis.

\bibliography{acl_latex}

\end{document}